\documentclass{article}

\usepackage{PRIMEarxiv}

\usepackage[utf8]{inputenc} % allow utf-8 input
\usepackage[T1]{fontenc}    % use 8-bit T1 fonts
\usepackage{hyperref}       % hyperlinks
\usepackage{url}            % simple URL typesetting
\usepackage{booktabs}       % professional-quality tables
\usepackage{amsfonts}       % blackboard math symbols
\usepackage{nicefrac}       % compact symbols for 1/2, etc.
\usepackage{microtype}      % microtypography
\usepackage{lipsum}
\usepackage{fancyhdr}       % header
\usepackage{graphicx}       % graphics
\graphicspath{{media/}}     % organize your images and other figures under media/ folder
\usepackage{caption}
\usepackage{subcaption}
\usepackage{algorithm2e}
\RestyleAlgo{ruled}
\usepackage{graphics}
\usepackage{makecell}

\usepackage[dvipsnames]{xcolor}

\newcommand{\ppm}{\,\scriptsize$\pm$}
\newsavebox\CBox
\def\textBF#1{\sbox\CBox{#1}\resizebox{\wd\CBox}{\ht\CBox}{\textbf{#1}}}

\newcommand{\CC}{C\nolinebreak\hspace{-.05em}\raisebox{.4ex}{\tiny\bf +}\nolinebreak\hspace{-.10em}\raisebox{.4ex}{\tiny\bf +}}
\def\CC{{\nolinebreak[4]\hspace{-.05em}\raisebox{.4ex}{\tiny\bf ++}}}

\newcommand{\mypar}[1]{\vspace{0.35\baselineskip}\noindent\textbf{#1}.\,}

\newcommand{\mr}[1]{\mathit{#1}}

\newcommand{\TTTpp}{TTT\CC}
\newcommand{\RecTTT}{\textit{ReC-TTT}}
\newcommand{\NCTTT}{NC-TTT}

\newcommand{\VisDA}{V\textsc{is}DA}

\newcommand{\trainToVal}{\mr{train}\!\to\!\mr{val}}
\newcommand{\trainToTest}{\mr{train}\!\to\!\mr{test}}
\title{ReC-TTT: Contrastive Feature Reconstruction for Test-Time Training}

\author{
  Marco Colussi~\textsuperscript{1} ~\thanks{Corresponding author: \texttt{marco.colussi@unimi.it}. \textit{\underline{Accepted for publication at WACV2025}}}
  \And
  Sergio Mascetti\textsuperscript{1} \\
  \And
  Jose Dolz\textsuperscript{2}\\
  \And
  Christian Desrosiers\textsuperscript{2}\\
  \and
  \textsuperscript{1} Università degli studi di Milano\\
  \and
  \textsuperscript{2}
  ÉTS Montréal \\
}

\begin{document}
\maketitle

\begin{abstract}
The remarkable progress in deep learning (DL) showcases outstanding results in various computer vision tasks. However, adaptation to real-time variations in data distributions remains an important challenge. Test-Time Training (TTT) was proposed as an effective solution to this issue, which increases the generalization ability of trained models by adding an auxiliary task at train time and then using its loss at test time to adapt the model.
Inspired by the recent achievements of contrastive representation learning in unsupervised tasks, we propose \RecTTT{}, a test-time training technique that can adapt a DL model to new unseen domains by generating discriminative views of the input data.
\RecTTT{} uses cross-reconstruction as an auxiliary task between a frozen encoder and two trainable encoders, taking advantage of a single shared decoder. This enables, at test time, to adapt the encoders to extract features that will be correctly reconstructed by the decoder that, in this phase, is frozen on the source domain. Experimental results show that \RecTTT{} achieves better results than other state-of-the-art techniques in most domain shift classification challenges.
    The code is available at: \url{https://github.com/warpcut/ReC-TTT}
\end{abstract}


\section{Introduction}
\label{sec:intro}

%Domain shift intro
The ability of deep learning (DL) models to generalize to new data is one of the main challenges that research in the field is facing. In the standard scenario, models are trained to learn patterns and relations on a source dataset, and performances are evaluated on a set of images not seen during training but extracted from the same distribution. Despite the impressive performance achieved by advanced models on various datasets, maintaining the assumption of domain invariance between source and target data proves to be impractical in many real-world scenarios. As a result, the limited robustness of DL models to distribution shifts remains a key obstacle to their use~\cite{torralba2011unbiased, miller2021accuracy}.

%Domain generalization vs domain adaptation.
As a solution to this challenge, two broad research branches have emerged: domain generalization (DG) and unsupervised domain adaptation (UDA). DG approaches aim at training more robust models with a native ability to generalize on various domains. The main limitation of these techniques is that they rely on the availability at train time of large amounts of data from different sources, which is often impractical~\cite{zhou2021domain, cha2021swad}. Moreover, these techniques may also underperform in domains very different from those considered during training.
 
On the other hand, UDA tries to achievehigher generalizability without anticipating potential distribution shifts but instead by adapting the model accordingly, either with test-time adaptation (TTA)~\cite{venkateswara2017deep, liang2020we} or test-time training (TTT)~\cite{sun2020test, hakim2023clust3}.
TTA methods perform adaptation without accessing the initial training step on the source data nor the labels of test samples. Our work focuses on the second scenario, TTT, which relaxes the very strict TTA setting and allows the design of the source training in a way that facilitates adaptation.    
%In-depth description of TTT + ReContrast --> RecTTT 
During training, TTT approaches follow a multi-task learning strategy where the model learns an auxiliary self-supervised, or unsupervised task, in addition to the main learning objective, sharing encoder features across tasks. 
%In this phase, a common feature extractor is trained.
Then, at test time, the model uses solely the auxiliary task to update the shared encoder with a loss that was already optimized for the source domain.

Our work leverages the concept of contrastive learning~\cite{chen2020simple} for improving test-time training. The idea of this powerful technique is to learn, from unlabeled data, general-purpose features that are similar in related samples and different in unrelated ones.
%making the features of related examples to be similar and those of unrelated ones to be different.
Previous work has shown the usefulness of contrastive learning in a variety of unsupervised and semi-supervised image tasks~\cite{chen2020simple, radford2021learning, he2020momentum}. Among others, \textit{ReContrast}~\cite{guo2024recontrast}, which inspired our approach, adopts feature reconstruction contrastive learning in \emph{unsupervised anomaly detection}, demonstrating a good transfer ability to various image domains compared to other unsupervised techniques. However, this recent approach has not been investigated for adapting models at test time.
%More recent techniques adopted the principles of contrastive learning in the fields of TTA and TTT, using contrastive loss as an auxiliary task~\cite{chen2022contrastive, zhang2022divide, huang2021model}.

This paper presents \RecTTT{} (Contrastive Feature Reconstruction for Test-Time Training), a test-time training approach designed for image classification that leverages techniques from the field of contrastive representation learning in a novel way.
The core idea of \RecTTT{} is to use a pre-trained frozen encoder to generate a discriminative feature representation of the input image.
This representation is then used as a positive pair in the learning of the auxiliary task. In particular, during the training phase, two encoders are trained in a supervised manner to classify the images and, at the same time, a decoder is trained to minimize the differences between the features extracted from the trainable encoders and the ones reconstructed from the frozen pre-trained encoder.
The intuition is that during test-time training, the now frozen decoder works as a guide to extract more meaningful information by the trainable encoders.

%Contributions
Our main contributions can be summarized as follows.
\begin{itemize}
\setlength\itemsep{.25em}
    \item %We propose the first test-time training approach using contrastive feature reconstruction as self-supervised task.
    We propose to use contrastive feature reconstruction as self-supervised task in TTT, which has been overlooked in the literature.
    \item We enhance vanilla contrastive feature reconstruction with an ensemble learning strategy where two classifiers are trained with different image augmentations to yield consistent predictions.   
    %\item %We thoroughly evaluate our approach, as well as several ablation variants, on various datasets with different types of distribution shifts. Our results demonstrate our approach's superior performance compared to recent TTA and TTA methods, reaching a new state-of-art.
    \item Comprehensive experiments on various datasets with different types of distribution shifts and relevant ablation studies demonstrate the superiority of our approach compared to recent TTA and TTT methods, yielding state-of-the-art performance.
    %\item Our approach has fewer parameters to tune compared to previous techniques, and these are evaluated through extensive ablation studies.
\end{itemize}

%\begin{comment}
In the next section, we provide an overview of existing TTA and TTT methods, emphasizing the novelty of our method presented in Section~\ref{sec:method}. Section~\ref{sec:exp} describes the experimental setting, reports the performances, and the ablation studies. Finally, our conclusions  and limitations are discussed in Section~\ref{sec:conclusion}.

\section{Related work}
\label{sec:related}

\mypar{Test-Time Adaptation (TTA)} In visual recognition tasks, distribution shifts between training and test data can greatly degrade performance~\cite{saenko2010adapting}. To overcome this issue, recent approaches proposed to dynamically adapt the models at test time to the new data. Unlike domain generalization~\cite{cha2021swad, zhou2022domain}, where the source model is robustly trained but fixed at test time, TTA allows updating the model for the target domain. 
TTA techniques do not have access to the source data or training (\textit{i.e.}, only the trained model is provided), and the adaptation happens only at test time. A variety of TTA methods have been proposed in recent years. Among others, in PTBN~\cite{nado2020evaluating}, the adaptation is carried out by updating the BatchNorm layer statistics using the test batch. Instead, TENT~\cite{wang2020tent} tries to minimize the entropy of the predictions for the test set. Finally, TIPI~\cite{nguyen2023tipi} proposes to identify transformations that can approximate the domain shift, and trains the model to be invariant to such transformations.

\mypar{Test-Time Training (TTT)} In contrast to TTA, TTT techniques %In TTT, unlike for TTA, 
have access to the source data during initial training (but not at test time), and a secondary self-supervised task is trained jointly with the main learning objective. %task. 
This learning paradigm was first introduced in TTT~\cite{sun2020test}, where the auxiliary task consists in recovering a random rotation of multiples of $90^{\circ}$. At test time, the adaptation is performed by updating only the parameters related to the secondary task.
TTT-MAE~\cite{gandelsman2022test} uses transformers as backbone for the supervised training, with a masked-autoencoders architecture trained as a self-supervised reconstruction task; at test time, the network is trained only to reconstruct the masked images, adapting the shared feature extractor. TTTFlow~\cite{osowiechi2023tttflow} adopts normalizing flows on top of a pre-trained network to map the features into a simple multivariate Gaussian distribution. At test time, the log-likelihood of this distribution is employed to adapt the model. 
ClusT3~\cite{hakim2023clust3} proposed an unsupervised clustering task that maximizes the Mutual Information between the features and the clustering assignment that should remain constant across different domains.

\mypar{Contrastive learning as auxiliary task} Contrastive learning as self-supervised task is gaining remarkable attention in domain adaptation research, thanks to its ability to learn robust representations. Among the various approaches based on this technique, 
AdaContrast~\cite{chen2022contrastive} takes advantage of both momentum contrastive learning and weak-strong consistency regularization for pseudo-label supervision. In DaC~\cite{zhang2022divide}, the test set is divided into source-like and domain-specific, applying two different strategies to the sub-sets using adaptive contrastive learning.
\TTTpp~\cite{liu2021ttt++} adopts a contrastive approach on top of a TTT framework, using two augmented versions of the same image as positive pairs and, as negative pairs, augmented versions of other images. This technique also leverages a batch-queue decoupling to regularize the adaptation with smaller batch sizes. 
More recently, \NCTTT{}~\cite{osowiechi2024nc} introduces a contrastive approach based on the synthetic generation of noisy feature maps.

The proposed \RecTTT{} method leverages contrastive feature reconstruction, whose ability to identify relevant domain-specific features was recently shown in anomaly detection~\cite{guo2024recontrast}. We extend this recent work by introducing the first TTT approach based on this technique. In addition to this new application setting, we enhance the vanilla contrastive feature reconstruction model of \cite{guo2024recontrast} in several important ways. First,
%Proposal
we introduce a classification task to the trainable encoder, learned jointly with the feature reconstruction. Second, instead of having a single trainable encoder as in \textit{ReContrast}\cite{guo2024recontrast}, our model includes a second one with its own classifier, processing an augmented version of the image. Finally, we introduce a regularization loss between the two classifiers, which promotes consistency across features extracted from the original and augmented data for an improved model generalization. 

Our novel architecture provides significant benefits compared to existing approaches for test-time training. While current TTT approaches based on contrastive learning rely solely on transformation invariance, our method also exploits the more general principle of reconstruction. However, unlike approaches such as TTT-MAE~\cite{gandelsman2022test} that measure reconstruction error on the output image, our method operates on the features of different network layers. This enables it to capture a broader spectrum of domain shifts (from pixel-level noise to image-level distortion). Last, our method improves upon these approaches via an ensemble learning strategy that combines the predictions of two independent classifiers.
%
%\jose{JD: Yes, in addition, it would be nice if we can motivate which benefits including contrastive feature reconstruction brings compared to existing contrastive learning in TTT.}
%

\section{Methodology}
\label{sec:method}

%\subsection{Problem formulation}

Our \RecTTT{} method addresses the problem of domain shift between a given training set, representing the source domain $\mathcal{S}=(X_S,Y_S)$, and a test set from a target domain $\mathcal{T}=(X_T,Y_T)$, where $X_S,X_T$ are spaces containing images and $Y_S,Y_T$ the spaces of corresponding labels. In this setting, we suppose that both domains have the same labels, i.e., $Y_S=Y_T$, but that images have a different conditional distribution, i.e., $p_S(x|y) \neq p_T(x|y)$ where $x\in X$ and $y\in Y$.

\begin{figure*}[ht]
  \centering
  \begin{subfigure}[b]{0.32\textwidth}
    \includegraphics[height=\textwidth]{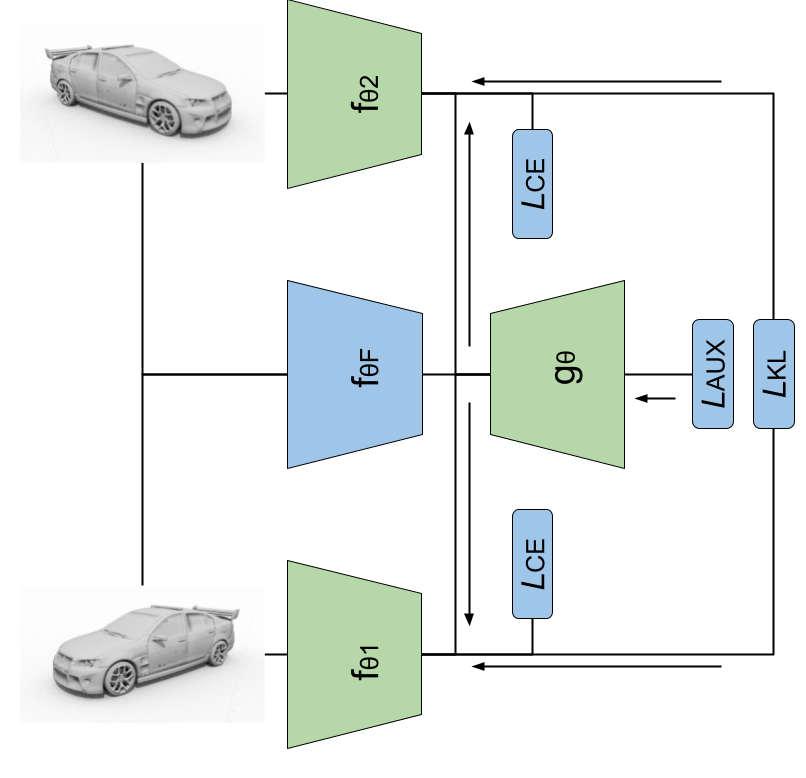}
    \caption{Train time}
    \label{fig:arch}
  \end{subfigure}
  \hspace{3mm}
  \hfill
  \begin{subfigure}[b]{0.32\textwidth}
    \includegraphics[height=\textwidth]{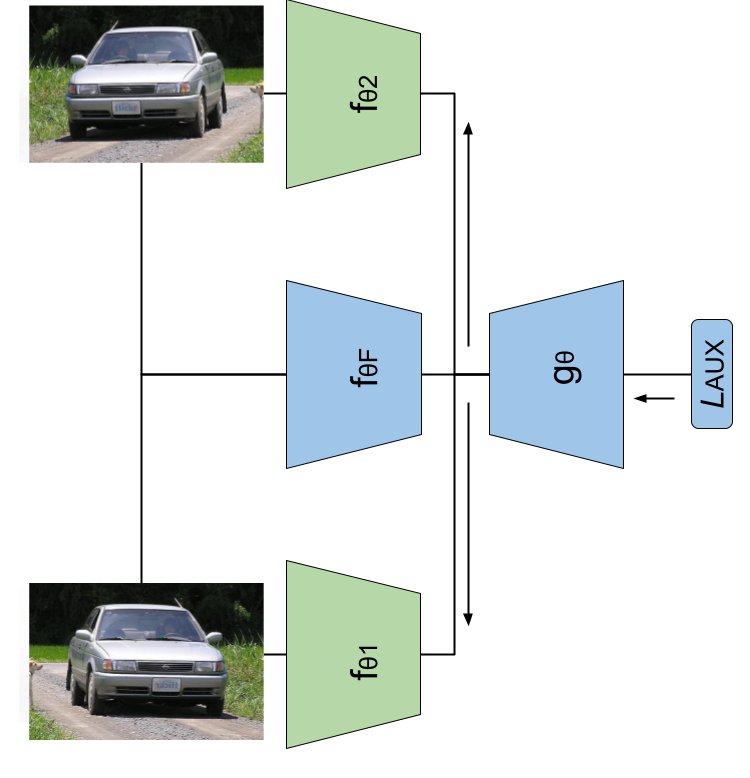}
    \caption{Test time}
    \label{fig:arch_t}
  \end{subfigure}
  \hfill
  \begin{subfigure}[b]{0.32\textwidth}
    \includegraphics[height=\textwidth]{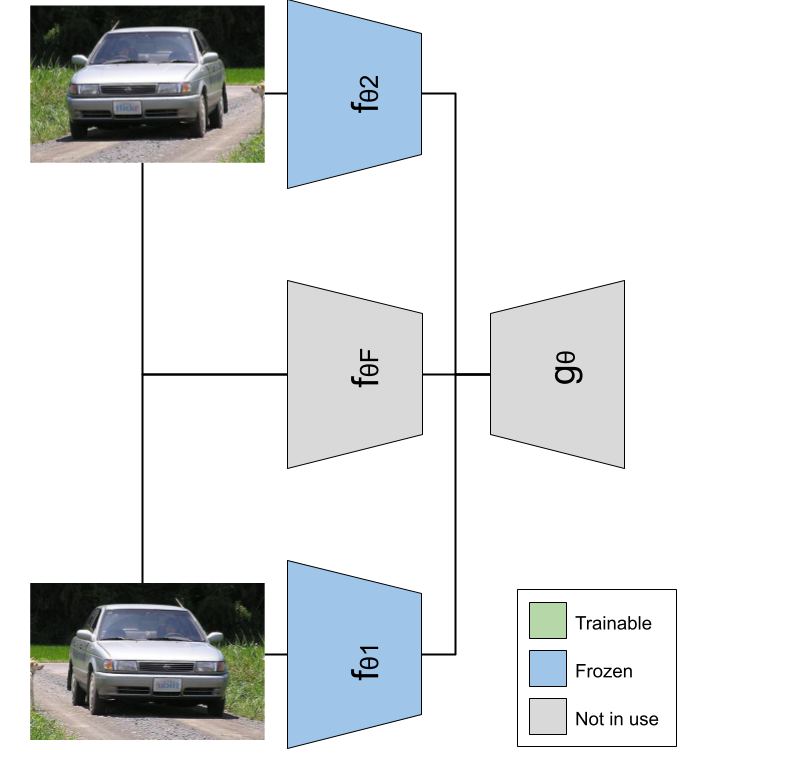}
    \caption{Inference time}
    \label{fig:arch_i}
  \end{subfigure}
    \caption{\textbf{Overview of our \RecTTT{} framework.} The directional flow of gradients is denoted by the symbol $\to$. $\mathcal{L}_{\mr{aux}}$ is our cross reconstruction loss, which computes the global similarity between the features of the encoders and the features reconstructed by the decoder, $\mathcal{L}_{\mr{CE}}$ is the cross-entropy between the predicted classes and the true labels, and $\mathcal{L}_{\mr{KL}}$ is the Kullback–Leibler divergence between the two predicted distributions. The trainable components of our architecture are depicted in \textbf{green}, whereas the frozen components are represented in \textbf{blue}. (\textbf{a}) illustrates the training phase, where both the encoders and the decoder are trainable. At test-time training (\textbf{b}), the decoder is frozen. Finally, (\textbf{c}) shows the inference time when the entire network is frozen; modules represented in \textbf{gray} are not needed in this phase.}
    \label{fig:recttt}
\end{figure*}

Figure \ref{fig:recttt} shows the overall framework of our method. The architecture employed in \RecTTT{} consists of two trainable encoders $f_{\theta1}$ and $f_{\theta2}$, a pre-trained frozen encoder $f_{\theta_F}$, and a decoder $g_{\theta}$ that takes in input the concatenated features extracted from the three encoders. As other TTT approaches, \RecTTT{} requires two steps. In the first step, our method has access to the source domain and the model is trained to learn a function mapping $X_S \to Y_S$ using a classification loss ($\mathcal{L}_{\mr{CE}}$) and an auxiliary loss ($\mathcal{L}_{\mr{aux}}$). The second step occurs at test time, where our method has only access to the unlabeled target set. In our case, $f_{\theta1}$ and $f_{\theta2}$ are updated using only the auxiliary function to learn the new mapping $X_T \to Y_T$. This partial update enables the model to learn the association in $\mathcal{T}$ while maintaining the knowledge acquired during training on $\mathcal{S}$. 

In the following sections, we detail the different components of our method. 

\subsection{Contrastive feature reconstruction}

Contrastive learning extracts meaningful representations by maximizing the agreement between the features of different views of the input data during training.
%In our implementation, this is achieved by using two encoders, one that is updated during training and the other that instead is frozen.
%The two encoders will therefore generate a domain-specific and a pre-trained domain representation.
%Check if this is what you mean
%The former generates a domain-specific domain representation, while the latter generates a pre-trained domain representation.
% I combine the previous two sentences:
In our framework, illustrated in Figure \ref{fig:recttt}, this is achieved using two separate encoders. The first one ($f_{\theta1}$) is updated during training, and hence generates a domain-specific domain representation, while the other ($f_{\theta F}$) is instead frozen and thus generates a domain representation based on a pre-trained network.

The extracted features are then combined into a bottleneck that resembles the last ResNet layer, and subsequently fed into a shared decoder ($g_{\theta}$) which has the opposite architecture of the encoders.
For a fair comparison with previous TTA and TTT works~\cite{hakim2023clust3, liu2021ttt++}, our method uses a ResNet50 backbone for the convolutional feature extractors.

\mypar{Learning objective}
The network is trained using global cosine-similarity~\cite{guo2024recontrast} between the features at different layers of the encoders and the features at the opposite level of the decoder. Specifically, the model is trained in a cross-reconstruction fashion where the decoder learns to reconstruct the features of the frozen encoder starting with the ones obtained by the trainable encoder, and vice versa, using the following loss:
\begin{equation}
\label{eq:globsim}
    \mathcal{L}_{\mr{aux}} \, = \, \sum_{\ell=1}^L 1-\frac{\big\langle sg(f_{E}^{\ell}), f_{D}^{\ell}\big\rangle}{sg(\|f_{E}^{\ell}\|_2)\,\|f_{D}^{\ell}\|_2}
\end{equation}
where $sg$ is the stop gradient operation~\cite{chen2021exploring} used to avoid propagating the gradient directly into the encoder, $f_{E}^{\ell}$ and $f_{D}^{\ell}$ represent the flattened features of the encoder and decoder respectively at the $\ell^{th}$ layer, and $\langle\cdot, \cdot\rangle$ is the dot product operation.

In the TTT framework, during training, the network is jointly trained on the two tasks: supervised classification and contrastive feature reconstruction combining the cross-entropy with the auxiliary loss described in Eq. (\ref{eq:globsim}), as follows:
\begin{equation}
\label{eq:train_simple}
    \mathcal{L}_{\mr{\mr{train}}} \, = \, \mathcal{L}_{\mr{CE}}+ \mathcal{L}_{aux}
\end{equation}

Experimental results (see ablation study reported in Section \ref{sec:cont} of supplementary material) show that the contrastive reconstruction loss outperforms arguably simpler solutions such as \textit{SimSiam}~\cite{chen2021exploring}.

\subsection{Encoder ensemble}

As shown in Fig.~\ref{fig:arch}, our \RecTTT{} method also leverages an ensemble learning strategy that integrates a secondary trainable encoder ($f_{\theta2}$) and classification predictor. This encoder takes as input an augmented version of the original image to learn diversified representations of the data and add robustness to the contrastive learning process. The same image is fed to the frozen encoder ($f_{\theta F}$) to generate two contrastive pairs, such that the representations extracted by the decoder should be invariant to the augmentation applied. To avoid introducing information that could artificially facilitate the adaptation to specific domain shifts (as noise), we selected a weak, domain-agnostic augmentation: horizontal flip.

\mypar{Learning objective} The model is trained with the loss of Eq.~(\ref{eq:train_simple}) applied to both encoders. Furthermore, we added a consistency loss between the two predictors, measuring their Kullback–Leibler (KL) divergence, to align the distributions predicted by the two encoders:
\begin{equation}
\label{eq:train_ens}
    \mathcal{L}_{\mr{\mr{train}}} \, = \, \mathcal{L}_{\mr{CE}}+ \mathcal{L}_{\mr{aux}} \, + \, \mathcal{L}_{\mr{KL}}
\end{equation}
Let $P$ and $Q$ be two discrete probability distributions over $k$ classes. The KL divergence is computed as 
\begin{equation}
D_{\mr{KL}}(P \, \| \, Q) \, = \, \sum_k p_k \log \left(\frac{p_k}{q_k}\right).
\end{equation}

\mypar{Adaptation} Algorithm~\ref{algo:TTT} describes how our method is used at test time for adapting the model to data from unseen domains. At this stage, we freeze the shared decoder and, for each test batch, reinitialize the weights of encoders. Afterward, since we have no access to labels for the supervised loss, the layers of the trainable encoders are updated using only Eq.~(\ref{eq:globsim}) for a total of $T$ iterations. For the final inference, the whole model is frozen, and we obtain the final classification by averaging the predictions of two independent encoders. 
As illustrated in Fig.~\ref{fig:arch_i}
to reduce computational complexity during inference, the architecture can be optimized by removing the frozen encoder and decoder, which are no longer necessary for generating predictions.

\begin{algorithm}[ht!]
\begin{small}
\caption{Test-Time Training Algorithm}
\KwData{Trained model parameters $\theta_0$, test set $X_{\mr{\mr{T}}}$}
\KwResult{Predicted labels $\hat{Y}$}
\BlankLine
\For{param $\in \theta_g$}{
    \textit{param.trainable} $\,\leftarrow\,$ False
}
\For{batch $\in X$}{
$\theta \leftarrow \theta_0$ \tcp*{Initialize weights}
\For{iter $t=1..T$}{
    Get layers features of batch samples $x$ using model with parameters $\theta$\;
    $\mathcal{L}_{\mr{aux}} \,\leftarrow\,$ Auxiliary loss between encoders and decoder\;
    $\nabla_\theta \mathcal{L} \leftarrow$ Compute gradient of $\mathcal{L}_{aux}$ with respect to $\theta_{t-1}$\;
    $\theta_t \, \leftarrow\, \theta_{t-1} - \alpha \,\nabla_\theta \mathcal{L}$ \tcp*{Update model parameters}
}

$\hat{y} \leftarrow$ Make prediction using $\theta_T$ on examples $x$\;
}
\Return{$\hat{Y}$}
\label{algo:TTT}
\end{small}
\end{algorithm}

\section{Experiments}
\label{sec:exp}

\subsection{Experimental setup}

Six publicly available datasets were selected for the evaluation. These datasets simulate various types of domain shift: image corruption, natural domain shift, and synthetic to real images.

\mypar{CIFAR-10C, CIFAR-100C and TinyImageNet-C\cite{hendrycks2019benchmarking}} These three datasets are composed of 15 different types of corruptions, from various types of noise and blur to weather and digital corruptions. The images present five levels of severity for each perturbation and all the experiments were conducted using only the most severe category (level 5). The datasets consist of $10,000$ test images labeled into $10$ classes for CIFAR-10C, $100$ classes for CIFAR-100C, and $200$ classes for TinyImageNet-C.

\mypar{CIFAR-10.1\cite{recht2019imagenet}} We also use the CIFAR-10.1 dataset to evaluate our model's ability to generalize to natural domain shift that takes place when images are re-collected after a certain time. The CIFAR-10.1 dataset is composed of $2,000$ images collected several years after the original CIFAR-10 dataset, with the same $10$ classes.

\mypar{\VisDA{}\cite{peng2018visda}} 
The Visual Domain Adaptation (VisDA) dataset was designed to pose a new challenge in domain adaptation: from synthetic images to real-world images. This dataset is composed of $152,397$ train images consisting of 2D renderings, $55,388$ validation images extracted from the COCO dataset, and $72,372$ YouTube video frames that compose the test set. All images are labeled into 12 different classes. We evaluated the model's ability to generalize from the training set to the validation set ($\trainToVal$) and from the training set to the test set ($\trainToTest$).

\mypar{Training protocol} Following previous work, our method employs Resnet50 as the backbone, using 32$\times$32 images for the CIFAR datasets, 64$\times$64 images for the TinyImageNet and 224$\times$224 for the \VisDA{} dataset. The backbone is pre-trained using ImageNet32~\cite{chrabaszcz2017downsampled} for the first one and ImageNet~\cite{deng2009imagenet} for the latter. %In agreement with previous approaches,
Following existing literature, all CIFAR models were trained for 300 epochs with SGD as optimizer, a batch size of 128, and an initial learning rate of $0.1$ with a multi-step scheduler, decreasing the learning rate by a factor of $0.1$ every 25 epochs. In contrast, for \VisDA{}, the model was trained for 100 epochs, a batch size of 64, and a learning rate of $0.001$ without a scheduler.

\mypar{Inference} At test time, the shared decoder is frozen, while the rest of the network is trained with SGD and a learning rate of $0.005$ for CIFAR datasets and $0.0001$ for \VisDA{}, using only the auxiliary loss. Similarly to previous approaches, we reset the weights after each batch, hence enabling the consequential processing of batches with different domain shifts.

The experiments were run on a Ubuntu server with an \texttt{NVIDIA A100} GPU, 42Gb of RAM, and an \texttt{AMD EPYC} 8-core CPU. The code is implemented in python3 with \texttt{PyTorch} $1.12.0$.%, is publicly available\footnote{\url{https://anonymous.4open.science/r/ReC-TTT-034D}}.

\subsection{Empirical results}
\subsubsection{Comparison with state of the art}

Our model was compared with seven recent approaches: ResNet50~\cite{he2016deep} as the baseline, trained with the same strategy as our method, but only on the supervised task, PTBN~\cite{nado2020evaluating}, TENT~\cite{wang2020tent}, \TTTpp~\cite{liu2021ttt++}\footnote{Due to reproducibility issue, \TTTpp{} results were extracted from the latest reported results~\cite{hakim2023clust3}.}, TIPI~\cite{nguyen2023tipi}, ClusT3~\cite{hakim2023clust3} and \NCTTT{}\cite{osowiechi2024nc}. For a fair comparison, all TTA methods were evaluated on the same pre-trained ResNet50, while TTT approaches were trained using the same ResNet50 base architecture and the same training strategy.

%\subsubsection{CIFAR-10 corruptions.}

\mypar{CIFAR-10 corruptions}
Table~\ref{tab:performance-comparison} shows the comparison on the CIFAR-10C dataset of the different state-of-the-art methods, the baseline, and our approach. It is noticeable that \RecTTT{} outperforms on average all previous methods, with a gain of $1.46\%$ on \TTTpp{} and $36.2\%$ on the baseline. Also, our method is the only one able to outperform the baseline for the natural domain shift (CIFAR 10.1, see last line in Table~\ref{tab:performance-comparison}).
As discussed in previous papers \cite{hakim2023clust3}, other techniques perform worse than the baseline possibly due to the small domain shift between CIFAR-10 and CIFAR-10.1.
%that is a challenge for the other methods. 
Instead, \RecTTT{} can capture this small domain shift thanks to more robust training, thus achieving better performances, 5.5\% higher than ClusT3 and around 3\% better than \NCTTT{}.
%thanks to the nature of a more robust training.
To be consistent with the other experiments, we report the performance with 20 adaptation iterations obtaining a gain of $0.27\%$ of AUROC score, while without adaptation our model achieves an even better AUROC of $90.18$ (see Section~\ref{sec:as2}). A common limitation of TTT methods is that they are subject to high variability. To investigate this aspect, we repeated the experiments three times with different random seeds (Table~\ref{tab:performance-comparison} reports the average among three runs).
The results show that the performance of \TTTpp{} can vary by $\pm5.05$, and those of ClusT3 by $\pm2.62$, whereas \RecTTT{} yields more consistent results with smaller variations (i.e., $\pm1.18$). %\textcolor{red}{Highlight that the differences in CIFAR 10.1 are (typically) much higher compared to the results observed in CIFAR-10c.}
%address this we show how our model is more consistent among three different runs with different seeds when compared with \TTTpp and ClusT3, where the variability can reach $\pm5.05$ for the first and $\pm2.62$ for the latter, \RecTTT{} shows a maximum of $\pm1.18$.

\begin{figure}
    \centering
    \includegraphics[width=\linewidth]{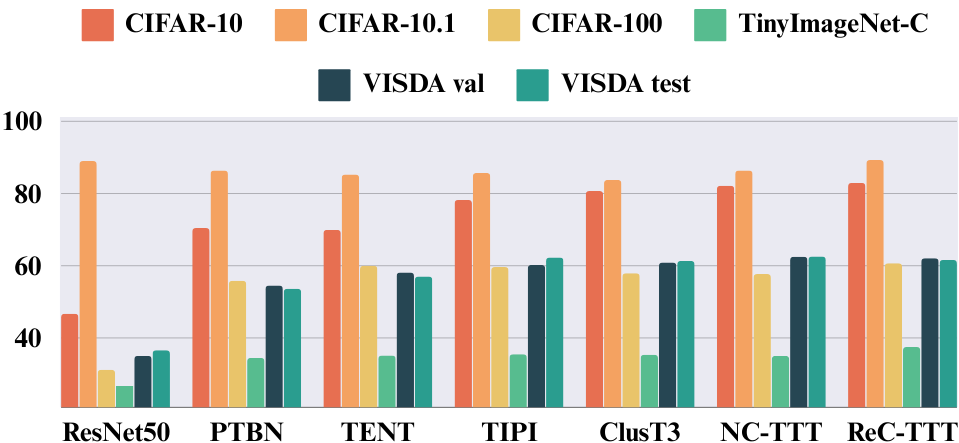}
    \caption{Quantitative results, compared to the state-of-the-art, on the CIFAR TinyImageNet-C and \VisDA{} datasets (\%). A detailed report for CIFAR-100, TinyImageNet and \VisDA{} is provided in supplementary material}.
    \label{fig:performance-comparison}
\end{figure}

%\subsubsection{CIFAR-100 corruptions.}
\mypar{CIFAR-100 and TinyImageNet-C corruptions} For the sake of the generability, the hyper-parameters used in the training of \RecTTT{} for CIFAR-100C and TinyImageNet-C are the same as those used for CIFAR-10C with the only difference that, for CIFAR-100C, the best results are obtained when all the layers are trainable.
Figure~\ref{fig:performance-comparison} shows that \RecTTT{} again outperforms the other techniques on both datasets, with a gain of $29.47\%$ when compared to the baseline for CIFAR-100C and a gain of $14.46\%$ for TinyImageNet-C. %Detailed results showing the performance with different corruptions are reported in the supplementary material.
This demonstrates the robustness of our method to adaptation settings involving a large number of classes.

%\subsubsection{\VisDA{}.}
\mypar{\VisDA{}} When training on \VisDA{}, \RecTTT{} achieves the best performance when all the layers of the encoders are trainable, and with 20 iterations of adaptation.
%Also \VisDA{} performed best when all the layers of the encoders were trainable, and with 20 iterations of adaptation.
Figure~\ref{fig:performance-comparison} also reports the results for ${\trainToVal}$ and ${\trainToTest}$ for \VisDA{}.
%Usually results are reported only for the $_{\mr{train}}\to _{\mr{val}}$ shift. In Table~\ref{tab:performance-comparison-visda} we also report the performances on the $_{\mr{train}}\to _{\mr{test}}$ problem.
\RecTTT{} performs better than all other approaches in ${\trainToVal}$ except \NCTTT{} while TIPI and \NCTTT{} show the best results for ${\trainToTest}$.
%results for  with \NCTTT{} having the highest performances among all the methods.
%\RecTTT{} demonstrates also in this case of being able to achieve better performances that previous approaches in $_{\mr{train}}\to _{\mr{val}}$. In $_{\mr{train}}\to _{\mr{test}}$ both TIPI and ClusT3 shows better performance, with TIPI beating our approach by $2.23\%$.
It is worth mentioning that ${\trainToVal}$ and ${\trainToTest}$ model the same synthetic-to-real domain shift, but using two sources of real images with different characteristics. For instance, the ratio of images for each class varies greatly, and images from the test set are obtained from video frame crops and may thus be blurry, etc. 

\begin{table*}[h]
\centering
\setlength{\tabcolsep}{4.5pt}
\resizebox{\textwidth}{!}{%
\begin{tabular}{l|ccccccccc}
\toprule
\bf Corruption Type & \bf ResNet50 & \bf PTBN & \makecell{\bf TENT \\[-1pt] {\footnotesize(ICLR20)}} & \makecell{\bf \TTTpp \\[-1pt] \footnotesize(NeurIPS21)} & \makecell{\bf TIPI \\[-1pt] \footnotesize(CVPR23)} & \makecell{\bf ClusT3 \\[-1pt]  \footnotesize(ICCV23)} & \makecell{\bf \NCTTT{} \\[-1pt] \footnotesize(CVPR24)} & \bf \RecTTT{} \\
\midrule
Gaussian Noise & 23.65 & 57.49 & 57.67 & 75.87\ppm5.05 & 71.90& \textBF{75.81}\ppm2.62 & 75.24 \ppm0.12 & 71.97\ppm1.18\\
Shot Noise & 27.68 & 61.07 & 60.82 & 77.18\ppm1.36 & 78.24 &77.32\ppm2.14 & \textBF{77.84} \ppm0.15 & 75.44\ppm1.02\\
Impulse Noise & 32.00 & 54.92 & 54.95  & \textBF{70.47}\ppm2.18 & 59.64 & 67.97\ppm2.78 & 68.77\ppm0.15& 69.28\ppm0.27\\
Defocus Blur & 38.73 & 82.23 & 81.39 & 86.02\ppm1.35 & 84.67 & 88.10\ppm0.20 & 88.22\ppm0.04& \textBF{89.56\ppm0.18}\\
Glass Blur & 36.49 & 53.91 & 53.45 & 69.98\ppm1.62 & 67.62 & 60.47\ppm1.72 & \textBF{70.19}\ppm0.18& 69.38\ppm0.73\\
Motion Blur & 49.85  & 78.38 & 78.13 & 85.93\ppm0.24 & 82.39 & 84.99\ppm0.49 & 86.82\ppm0.10 & \textBF{88.94\ppm0.03}\\
Zoom Blur & 44.58 & 80.87 & 80.56 & 88.88\ppm0.95 & 85.01 & 86.76\ppm0.29 & 88.36\ppm0.10& \textBF{89.65\ppm0.27}\\
Snow & 65.39 & 72.06 & 71.46 & 82.24\ppm1.69 & 80.68 & 81.46\ppm0.39 & 84.42\ppm0.07& \textBF{86.75\ppm0.44}\\
Frost & 48.55 & 68.68 & 68.81 & 82.74\ppm1.63 & 82.12 & 80.73 \ppm1.25 & 84.80\ppm0.06 & \textBF{86.83\ppm0.59}\\
Fog & 58.81 & 76.32 & 75.94 & 84.16\ppm0.28 & 76.05 & 82.52\ppm0.25 & 86.81\ppm0.12& \textBF{88.87\ppm0.33}\\
Brightness & 84.72 & 85.38 & 84.87 & 89.97\ppm1.20 & 88.96 & 91.52\ppm0.24 & 92.52\ppm0.04 & \textBF{94.03\ppm0.24}\\
Contrast & 25.38 & 81.27 & 80.65 & 86.60\ppm1.39 & 76.49 & 82.59\ppm0.92 & 87.84\ppm0.11 & \textBF{89.56\ppm0.48}\\
Elastic Transform & 60.90 & 67.76 & 67.21 & 78.46\ppm1.83 & 77.25 & 80.04\ppm0.35 & 80.23\ppm0.06& \textBF{81.66\ppm0.32}\\
Pixelate & 39.25 & 69.59 & 69.22 & 82.53\ppm2.01 & 82.67 & 81.69\ppm0.58 & 81.93\ppm0.22 & \textBF{82.13\ppm0.34}\\
JPEG Compression & 64.96 & 66.50 & 66.17 & 81.76\ppm1.58 & 79.39 & \textBF{81.58}\ppm1.18 & 78.49 \ppm0.09 & 79.69\ppm0.12\\
\midrule
\textBF{Average} & 46.73 & 70.43 & 69.93 & 81.46 & 78.21 & 80.67 & 82.17 & \textBF{82.92}\\
\midrule
\textBF{CIFAR 10.1} & 89.00 & 86.40 & 85.30 & 88.03 & 85.70 & 83.77 & 86.40 & \textBF{89.27}\\
%\midrule
%\textBF{CIFAR 100-C} & 31.19 & 55.83 & 60.44 & -- & 59.63 & 57.89 & \textBF{60.66} \\
\bottomrule
\end{tabular}
} % end of \resizebox
\caption{Performance comparison with state-of-the-art on CIFAR-10C and CIFAR10.1 (\%).}
\label{tab:performance-comparison}
\end{table*}

\subsubsection{Visualization of the adaptation}
\begin{figure*}[ht]
  \centering
  \begin{subfigure}[]{0.23\textwidth}
    \includegraphics[width=\textwidth]{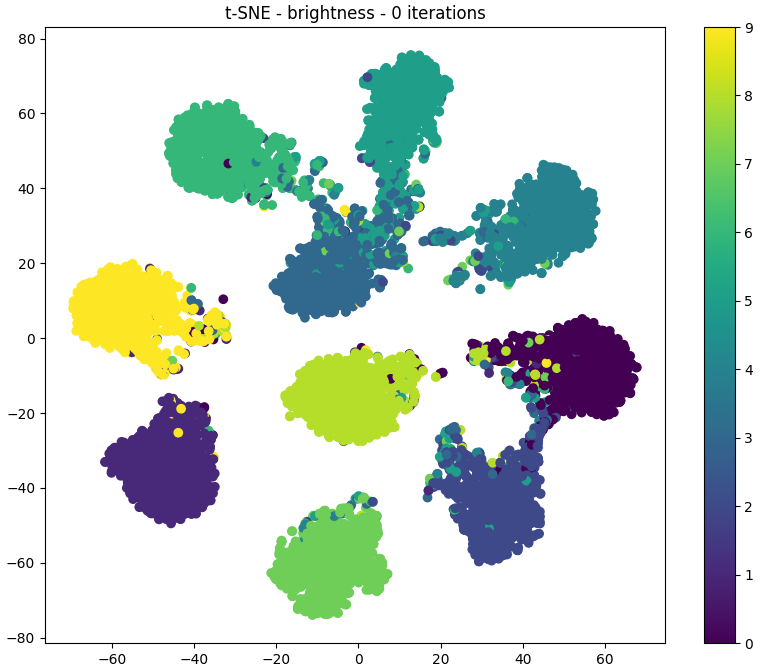}
    \caption{0 iteration}
    \label{fig:b_0}
  \end{subfigure}
  \hfill
  \begin{subfigure}[]{0.23\textwidth}
    \includegraphics[width=\textwidth]{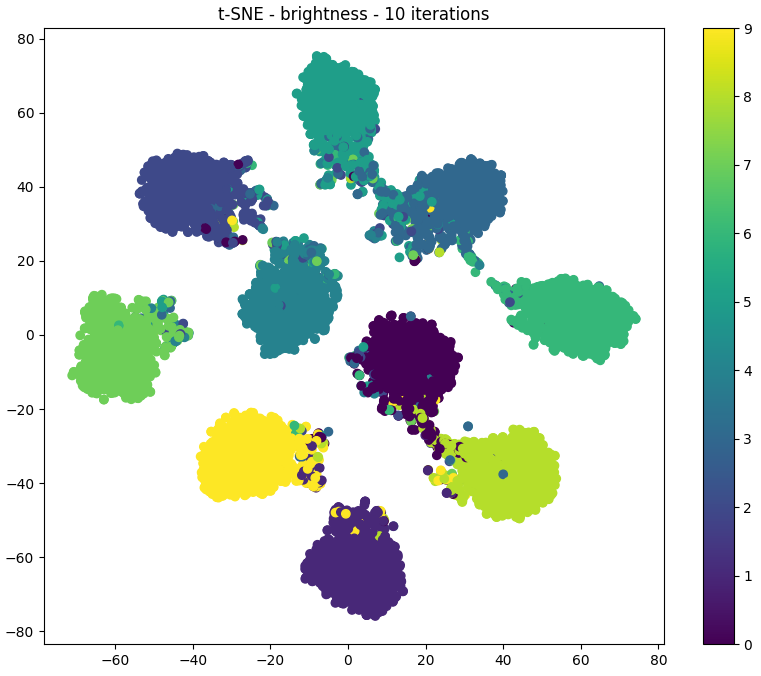}
    \caption{10 iterations}
    \label{fig:b_10}
  \end{subfigure}
  \hfill
  \begin{subfigure}[]{0.23\textwidth}
    \includegraphics[width=\textwidth]{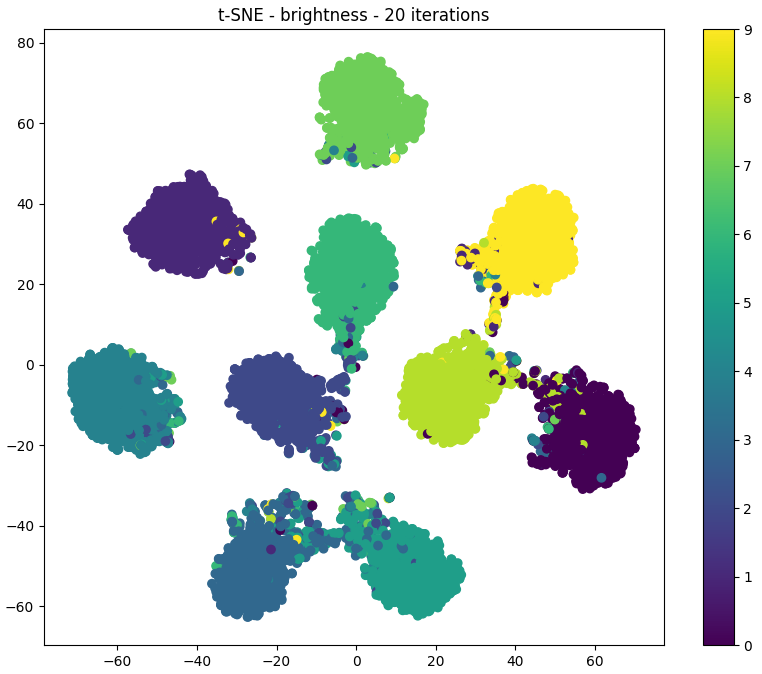}
    \caption{20 iterations}
    \label{fig:b_20}
  \end{subfigure}
  \hfill
  \begin{subfigure}[]{0.23\textwidth}
    \includegraphics[width=\textwidth]{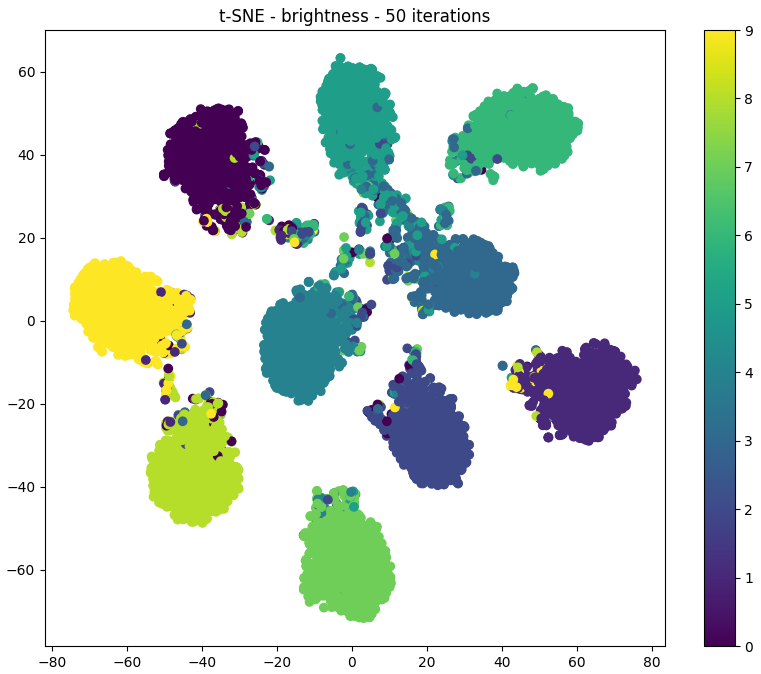}
    \caption{50 iterations}
    \label{fig:b_50}
  \end{subfigure}
  \begin{subfigure}[]{0.23\textwidth}
    \includegraphics[width=\textwidth]{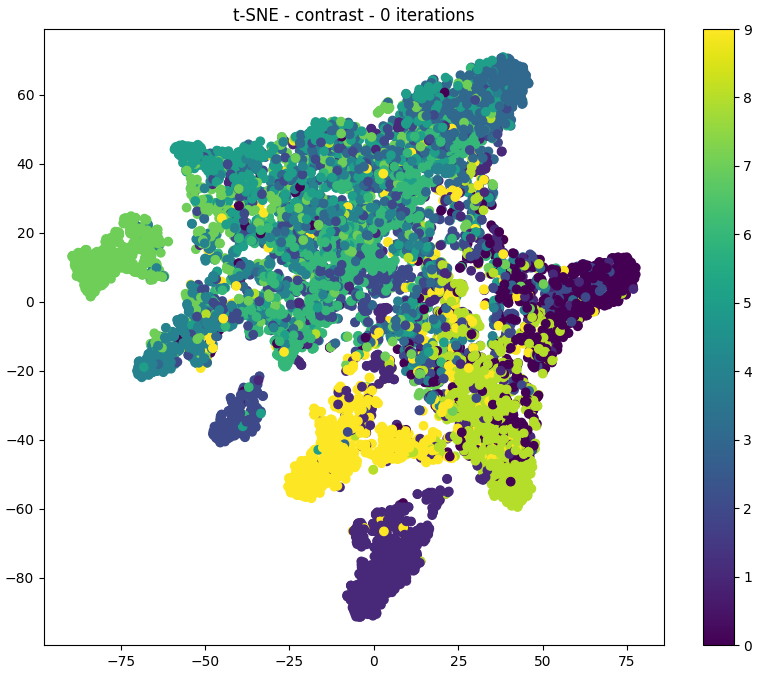}
    \caption{0 iteration}
    \label{fig:c_0}
  \end{subfigure}
  \hfill
  \begin{subfigure}[]{0.23\textwidth}
    \includegraphics[width=\textwidth]{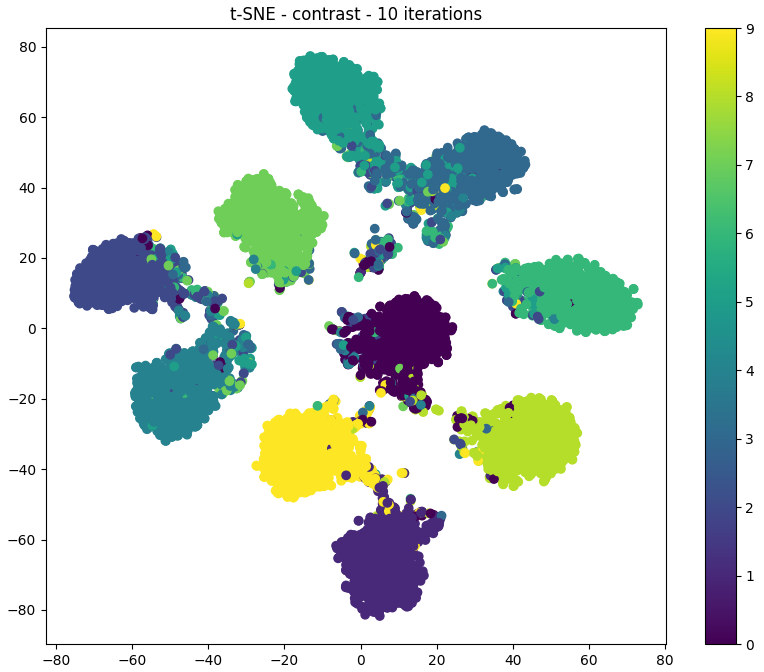}
    \caption{10 iterations}
    \label{fig:c_10}
  \end{subfigure}
  \hfill
  \begin{subfigure}[]{0.23\textwidth}
    \includegraphics[width=\textwidth]{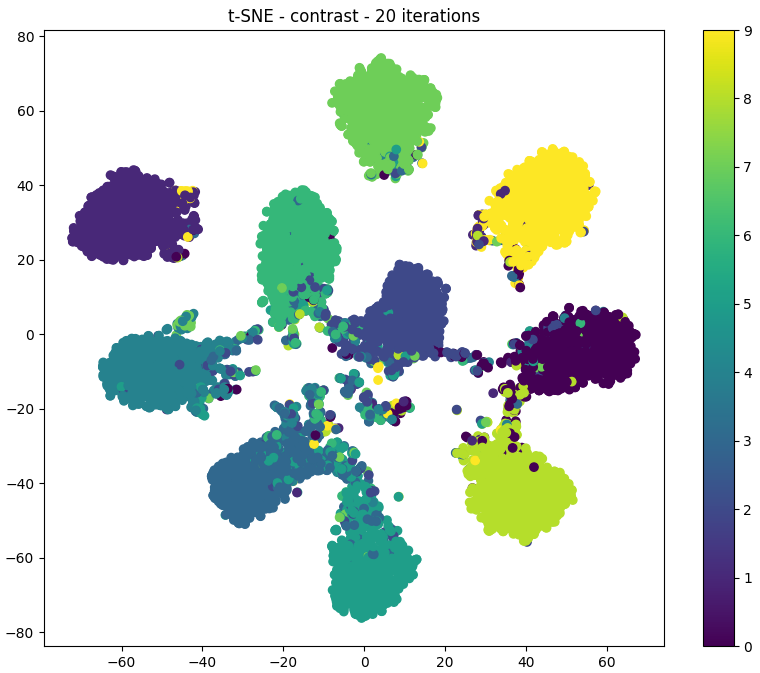}
    \caption{20 iterations}
    \label{fig:c_20}
  \end{subfigure}
  \hfill
  \begin{subfigure}[]{0.23\textwidth}
    \includegraphics[width=\textwidth]{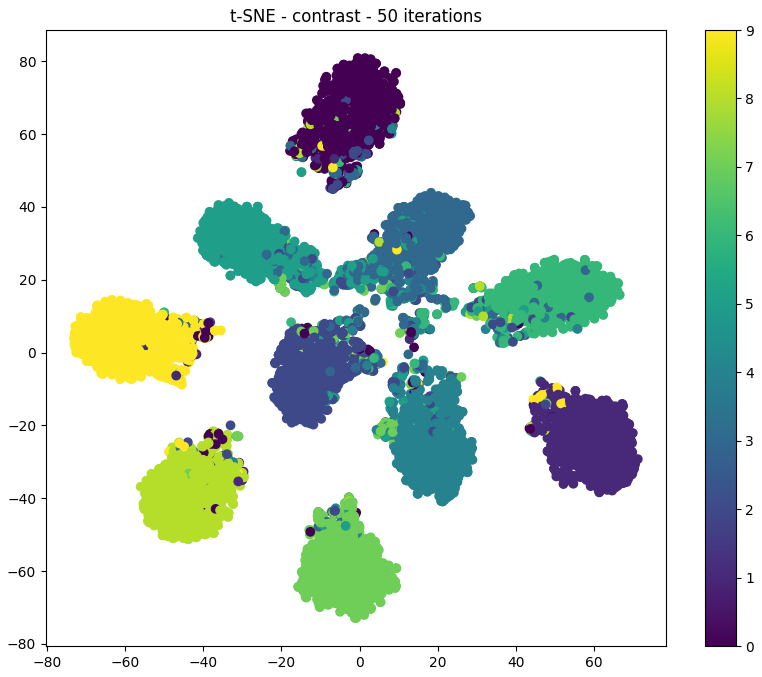}
    \caption{50 iterations}
    \label{fig:c_50}
  \end{subfigure}
  \caption{t-SNE plot of features after different adaptation iterations (0, 10, 20, 50) for the \textit{Brightness} (top row) and \textit{Contrast} (bottom row) corruptions of CIFAR-10C. The adaptation at test time helps separate the features of examples from the same class (represented by color).}
  \label{fig:tsne}
\end{figure*}

To better understand the effect of adaptation, we consider Figure~\ref{fig:tsne} showing the t-SNE plots of the test features before adaptation and after different numbers of iterations for two corruptions types: Brightness and Contrast.
%To better understand the effect of adaptation, in Figure~\ref{fig:tsne} we show the t-SNE plots of the test features before adaptation and after different numbers of iterations for the less (Brightness) and the most (Contrast) impactful adaptations.
In the top row (Brightness), we can observe that \RecTTT{} obtains a good separation of the features for the different classes also without iterations (AUROC of $92.31$). Successive iterations further separate the clusters but have a marginal impact on performance (AUROC of $94.03$ at iteration $20$).
%In the first row is noticeable how the network is able to obtain a good separation of the features for the different classes, while with as adaptation iterations increase, the clusters tend to be better separated.
The results are different in the bottom row (Contrast). In this case, without adaptation, most features overlap, without a clear separation, and, indeed, \RecTTT{} reaches an AUROC of $48.14$. With successive iterations, the cluster separation improves (AUROC of $89.56$ at iteration $20$) thus demonstrating the effectiveness of our adaptation technique on the extracted features.
On the other hand, for a limited number of samples that are wrongly classified before the adaptation, the distance of these samples to the true class increases with the number of iterations.
%Although this demonstrates , we can also notice that some examples that are wrongly classified before the adaptation are pushed further away from their true class during adaptation.
%In the second row, where the model without adaptation reaches an AUROC of $48.14$, the majority of the features are overlapped in a big cluster, and after the various number of iterations the clusters are more and more separated. Although this demonstrates the effectiveness of our adaptation on the extracted features, and hence show a desirable effect, we can also notice that with the increase of iterations some examples wrongly classified are pushed further from their true class.

\subsubsection{Robustenss to smaller batch sizes}
As demonstrated in \cite{nguyen2023tipi}, most domain adaptation approaches suffer from the need for large batch sizes to achieve competitive results. Most methods are usually evaluated with batches that have a size of $128$ or more.
This is a limitation in the application in which it is not possible to collect large batches before computing the inference.
%This emerges as a limitation when considering the applicability of TTT and TTA techniques in real-world scenarios, since the amount of data available in a single batch is not known.
For this reason, we compared the performances using different batch sizes (8, 32, 64, 128) for the CIFAR-10C dataset.
Table~\ref{tab:batch-comparison} reports the results of this study showing that most of the SOTA approaches lose up to $7\%$ of AUROC when the number of samples is lower than the number of available classes (a performance degradation is also reported for TIPI in \cite{nguyen2023tipi}). In contrast, the performance loss of \RecTTT{} is less than $2\%$ even with the smallest batch size. %\textbf{TODO: what about TIPI?}

\begin{table}[h]
\centering
\setlength{\tabcolsep}{5pt}
\small
\resizebox{0.7\linewidth}{!}{%
\begin{tabular}{c|c| c|c|c|c}
\toprule
Batch size & PTBN & TENT %& TIPI 
& ClusT3 & \NCTTT & \RecTTT{} \\
\midrule
8  & 61.97 & 62.11 % & \textbf{82.29} 
& 73.73& 79.75 & \textBF{81.68}   \\
32 & 68.46 & 68.46 % & 81.77 
& 80.14 & 81.80& \textBF{82.54}   \\
64 & 69.45 & 69.54 % & 80.09 
& 80.38 & 82.01 & \textBF{82.84}   \\
128 & 70.43 & 70.09 %& 78.21 
& 80.67 & 82.43 & \textBF{82.92}   \\
\bottomrule
\end{tabular}
}
\caption{\textbf{Robustness to the batch size.} Qualitative performance of different approaches on CIFAR-10C (\%) for different bacth sizes used during training.}
\label{tab:batch-comparison}
\end{table}

\subsubsection{Which layers to adapt?}
Previous studies suggest that the selection of layers that are updated by the auxiliary task at test time can affect performance~\cite{liu2021ttt++, lee2022surgical, hakim2023clust3}.
Table~\ref{tab:layer-comparison} reports the results of the adaptation of different layers on CIFAR-10C, having the best performance when updating only the first three ResNet blocks. While the difference in performance between three or four trainable layers is negligible ($+0.17\%$), adapting the first two layers yields a reduction in performance of $2.3\%$, and using the first layer only results in a drop of performance of $18.13\%$.
Differently, for CIFAR-100C and \VisDA{}, we obtained the best results by updating all four layers of the trainable encoders. This can be explained by the greater challenge posed by these datasets, i.e., the larger number of classes for CIFAR-100C and the harder synth-to-real domain shift for \VisDA{}, requiring adaptation of features in deeper layers.

\begin{table}[h]
\centering
\setlength{\tabcolsep}{5pt}
\resizebox{0.7\linewidth}{!}{%
\begin{tabular}{c|c|c|c|c}
\toprule
Trainable layers & Impulse Noise & Brightness & Pixelate & Average \\
\midrule
1 layer\phantom{s}  & 16.12 & 93.26 & 34.83 & 64.79 \\
2 layers & 61.86& 94.00 & 76.46 & 80.62 \\
3 layers & 69.25 & 94.06 & 82.03& \textBF{82.92}   \\
4 layers & 69.56 & 93.78 & 81.57 & 82.75   \\
\bottomrule
\end{tabular}
}
\caption{\textbf{On the impact of the training different layers.} Performance comparison of training different layers of our approach on CIFAR-10C (\%).}
\label{tab:layer-comparison}
\end{table}

\subsubsection{Number of adaptation iterations}
\label{sec:as2}
\begin{figure}[ht!]
  \centering
    \includegraphics[width=\linewidth]{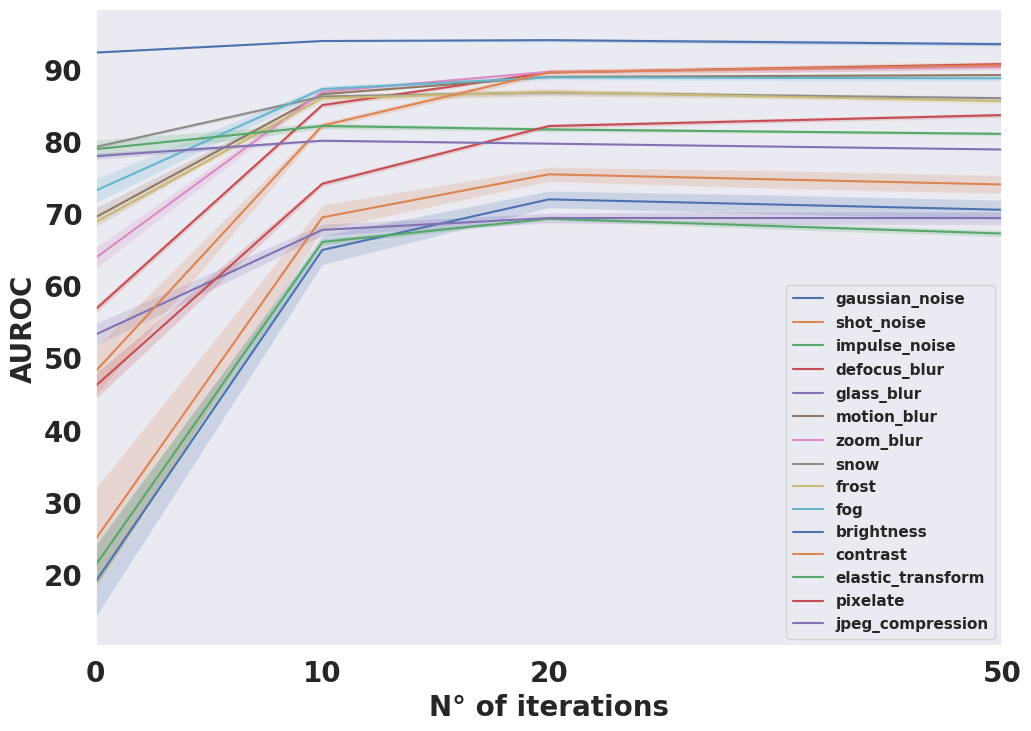}
    \caption{\textbf{How many iterations are needed for adaptation?} Performance (AUROC) obtained by our method with different number of adaptation iterations on CIFAR-10C. For most corruption types, our method provides a significant boost within few iterations and remains stable when the number of iterations is increased. }
    \label{fig:iterations}
\end{figure}

Another important aspect of test-time adaptation is the number of iterations needed at test time to obtain the best results.
In line with previous studies~\cite{osowiechi2023tttflow, hakim2023clust3} Figure~\ref{fig:iterations} shows, for the corruption types of CIFAR-10C, that the best results are obtained after 20 iterations. Successive iterations do not yield better results, on average. The same image for CIFAR-100 corruptions can be found in supplementary material.
%remaining constant after that.
The same finding emerges from the other datasets with the only exception of CIFAR-10.1 where, as per previous experiments \cite{osowiechi2023tttflow, hakim2023clust3, osowiechi2024nc}, adaptation tends to degrade the performances ($90.18\%$ of AUROC without adaptation, $89.27\%$ of AUROC after $20$ iterations).

\subsubsection{Impact of removing the second trainable encoder}
We evaluated the effectiveness of \RecTTT{}'s ensemble learning strategy,
which employs two trainable encoders,
by comparing it with the base architecture with a single encoder.
%As our architecture uses an ensamble of classifiers and encoders, we evaluate the effectiveness of this addition to the base ReContrast architecture.
Table~\ref{tab:arch-comparison} shows the performance for three different corruptions and the average among all 15 corruptions available in CIFAR-10C. We observe that using two encoders performs better than having a single one in all cases. This demonstrates the effectiveness of our ensemble learning approach to stabilize training and provide a more robust prediction. To address the increased computational complexity introduced by our model, we investigated the impact of using only one encoder during inference when constrained by performance requirements. While this approach results in a slight reduction in performance, it still yields better results when compared to training the model without the ensemble architecture, remaining competitive with the other SOTA approaches.
%In contrast, removing the second trainable encoder results in the loss of $1.75\%$ AUROC.
%In Table~\ref{tab:arch-comparison} we show the performances for three different perturbations and the average among all the 15 available in CIFAR10-C. This demonstrates the effectiveness of having a more complex architecture to stabilize the training and encourage a more robust representation learning, loosing $1.75\%$ when removing the second trainable encoder.

%\rev{For comparison, we also applied an ensembling strategy to NC-TTT. However, the effect is slightly different, as this approach applies ensembling solely during the inference stage, whereas in our approach, the secondary ensemble actively participates during both the training and adaptation stages. The results are reported in the last line of Table~\ref{tab:arch-comparison}, as we can observe \textbf{TODO.. waiting for the results}}

\begin{table}[t!]
\centering
\setlength{\tabcolsep}{5pt}
\resizebox{0.7\linewidth}{!}{%
\begin{tabular}{l|c|c|c|c}
\toprule  
         & Impulse Noise & Brightness & Pixelate & Average \\
\midrule
One encoder & 65.19 & 93.17 & 80.36 & 81.07\\
\midrule
Two encoders & 69.28 & 94.03 & 82.13 & \textBF{82.82}\\
\makecell{One encoder \\ (Inference)} & 67.54 &  93.31 & 80.89 & 81.59\\
%\midrule
%\makecell{NC-TTT \\ (Two encoders)} &  &  &  & \\
\bottomrule
\end{tabular}
}
\caption{\textbf{Using one \textit{vs.} two encoders.} Qualitative results on different configurations of our approach, on CIFAR-10C (\%).}
\label{tab:arch-comparison}
\end{table}

\section{Conclusion}
\label{sec:conclusion}

Our work addressed the problem of domain shift between training and test data under the Test-Time Training framework. We presented \RecTTT{}, a novel TTT approach based on contrastive feature reconstruction that can efficiently and effectively adapt the model to new unseen domains at test-time. Through a series of extensive experiments, we demonstrated that our model outperforms other state-of-the-art approaches across diverse datasets subject to different distribution shifts. An important limitation of previous approaches is their need for large batches of test samples to correctly adapt the model. Our results show that \RecTTT{} is more robust to this factor, even when the number of classes is greater than the number of available samples. Furthermore, we highlight the robustness of our method against training variability, typically observed in current TTT approaches. Another key advantage of our approach is that it requires tuning few hyper-parameters at test-time, specifically, the layers to adapt, the learning rate, and the number of adaptation iterations. These results underscore the potential of our method in enhancing the applicability of TTT in various domains under challenging conditions.

\subsection{Limitations and future work} 
While the results presented in this work show how to tackle the domain-shift problem with test-time training effectively, we have to consider certain limitations: as in all TTT approaches, one key limitation is the need for the source data to train the model. Moreover, \RecTTT{} was evaluated using only ResNet50 as backbone architecture, however employing different feature extractors might potentially yield a better performance. The architecture introduces multiple encoders that might affect the computational and memory requirements during training and adaptation, however, during inference, these additional components are not required, mitigating the resource demands. Last, although we validated our method on three classic benchmarks, further evaluation should be performed on different datasets representing different domain shifts.

\section*{Acknowledgments}
This project was partially supported by TEMPO – Tight control of treatment efficacy with tElemedicine for an improved Management of Patients with hemOphilia project, funded by the Italian Ministry of University and Research, Progetti di Ricerca di Rilevante Interesse Nazionale (PRIN) Bando 2022 - grant [2022PKTW2B] and MUSA (Multilayered Urban Sustainability Action) project funded by the NextGeneration EU program. We also thank \textit{Compute Canada}.

\newpage
\appendix
\begin{center}
    \textbf{\large ReC-TTT: Contrastive Feature Reconstruction for Test-Time Training  \\Supplementary Material} % Centered title for the appendix
\end{center}

\section{Extended results}

For CIFAR-100C, TinyImageNet-C and \VisDA{} our model was compared with the same state-of-the-art approaches except \TTTpp{} where the results were not reproducible nor available: ResNet50~\cite{he2016deep}, PTBN~\cite{nado2020evaluating}, TENT~\cite{wang2020tent}, TIPI~\cite{nguyen2023tipi}, ClusT3~\cite{hakim2023clust3} and \NCTTT{}\cite{osowiechi2024nc}. As per previous experiments TTA methods were evaluated on the same pre-trained ResNet50, while TTT approaches were trained using the same ResNet50 base architecture and the same training strategy.

\subsection{VisDA}

Table~\ref{tab:performance-comparison-visda} reports the detailed results on the VisDA dataset. \RecTTT{} outperforms most approaches on average, with a notable increase compared to the ResNet50 baseline without adaptation ($+25.81$). On $\trainToVal$ and $\trainToTest$, \NCTTT{} performs better than \RecTTT{}  ($\approx\!+1\%$ on average). %However, results show how recent TTT methods are more solid on complex datasets such as \VisDA{} compared to Source, PTBN, TENT, which are more competitive on the CIFAR datasets. These results can be because the reconstruction task is capable of capturing better general features while more basic approaches struggle to identify the more subtle domain shift.
Moreover, the results demonstrate that TTT methods show greater robustness on complex datasets, such as \VisDA{}, compared to methods like Source, PTBN, and TENT, which are more competitive on the CIFAR datasets. This performance difference may be attributed to the reconstruction task's ability to capture more generalizable features, while simpler approaches struggle to detect more subtle domain shifts.

\begin{table}[h]
\centering
\caption{Performance comparison with state-of-the-art on \VisDA{} dataset (\%).}
\label{tab:performance-comparison-visda}
\small
\setlength{\tabcolsep}{4.5pt}
%\resizebox{\linewidth}{!}{%
\begin{tabular}{l|c|c|c}
\toprule
 & \VisDA{}\,${\trainToVal}$ & \VisDA{}\,${\trainToTest}$ & Average\\
 \midrule
 \bf ResNet50 & 35.01 & 36.58 & 35.80\\
 \bf PTBN & 54.53& 53.63 & 54.08\\
 \bf TENT & 58.13& 57.04 & 57.59\\
 %\midrule
 %\bf \TTTpp{} & 60.42 & -- & --\\
 \bf TIPI & 60.22 & 62.26 & 61.24\\
 \bf ClusT3 & 60.89 & 61.33 & 61.11\\
 \bf \NCTTT{} & \textBF{62.49} & \textBF{62.57} & \textbf{62.53}\\
 \midrule
 \bf \RecTTT{}  & 62.06 & 61.12 & 61.59\\ %& 61.08 & 59.98\\ % & 62.08 & 61.14
\bottomrule
\end{tabular}
%} % end of \resizebox
\end{table}

\subsection{CIFAR-100C}

Table~\ref{tab:performance-comparison-100} shows in detail the results and the comparison with state-of-the-art approaches on all the perturbations of CIFAR-100C. \RecTTT{} the best results, demonstrating a 30\% increase in AUROC after adaptation compared to the baseline. This improvement surpasses the most recent state-of-the-art approaches as ClusT3 and \NCTTT{} by 3\%.

\begin{table*}[h]
\centering
\setlength{\tabcolsep}{4.5pt}
%\resizebox{\textwidth}{!}{%
\begin{tabular}{l|ccccccccc}
\toprule
\bf Corruption Type & \bf ResNet50 & \bf PTBN & \bf TENT & \bf TIPI & \bf ClusT3 & \bf \NCTTT & \bf \RecTTT{} \\
\midrule
Gaussian Noise & 13.23 & 42.30 & 51.35 & 48.88 & \textBF{52.79} & 46.03 & 48.12 \\
Shot Noise & 15.46 & 43.30 & 52.63 & 50.61 & \textBF{52.91} & 47.04 & 50.43 \\
Impulse Noise & 7.89 & 37.41 & 45.39 & 43.80 & \textBF{45.54} & 41.53 & 45.29 \\
Defocus Blur & 27.36 & 67.46 & 69.44 & 68.72 & 66.66 & 67.00 & \textBF{71.21} \\
Glass Blur & 21.18 & 46.44 & \textBF{51.01} & 50.93 & 50.76 & 48.08 &  49.94 \\
Motion Blur & 38.18 & 64.21 & 67.27 & 66.63 & 62.92 & 64.31& \textBF{68.86} \\
Zoom Blur & 32.81 & 66.68 & 69.33 & 68.84 & 65.42 & 66.24 & \textBF{69.91} \\
Snow & 44.85 & 55.52 & \textBF{60.47} & 59.51 & 56.65 & 58.70 & 60.21 \\
Frost & 31.56 & 54.76 & 58.35 & 57.90 & 56.91 & 58.55 & \textBF{60.16} \\
Fog & 32.79 & 56.77 & \textBF{62.29} & 61.12 & 53.95 & 57.73 & 62.22 \\
Brightness & 66.13 & 68.97 & 71.40 & 71.00 & 66.78 & 71.36 & \textBF{73.47} \\
Contrast & 11.87 & 63.47 & 65.63 & 65.17 & 56.46 & 61.53 & \textBF{67.06} \\
Elastic Transform & 48.87 & 57.93 & 60.07 & 59.94 & 59.07 & 60.25 & \textBF{62.37} \\
Pixelate & 26.70 & 59.75 & \textBF{64.06} & 63.56 & 62.26 & 61.17 & 63.61 \\
JPEG Compression & 48.88 & 52.45 & 57.84 & 57.79 & \textBF{59.34} & 55.69 & 57.05 \\
\midrule
\textBF{Average} & 31.19 & 55.83 & 60.44 & 59.63 & 57.89 & 57.68 & \textBF{60.66} \\
\bottomrule
\end{tabular}
%} % end of \resizebox
\caption{Performance comparison with state-of-the-art on CIFAR-100C perturbations (\%).}
\label{tab:performance-comparison-100}
\end{table*}

\subsubsection{Number of adaptation iterations}
\begin{figure*}[ht]
  \centering
    \includegraphics[width=.9\textwidth]{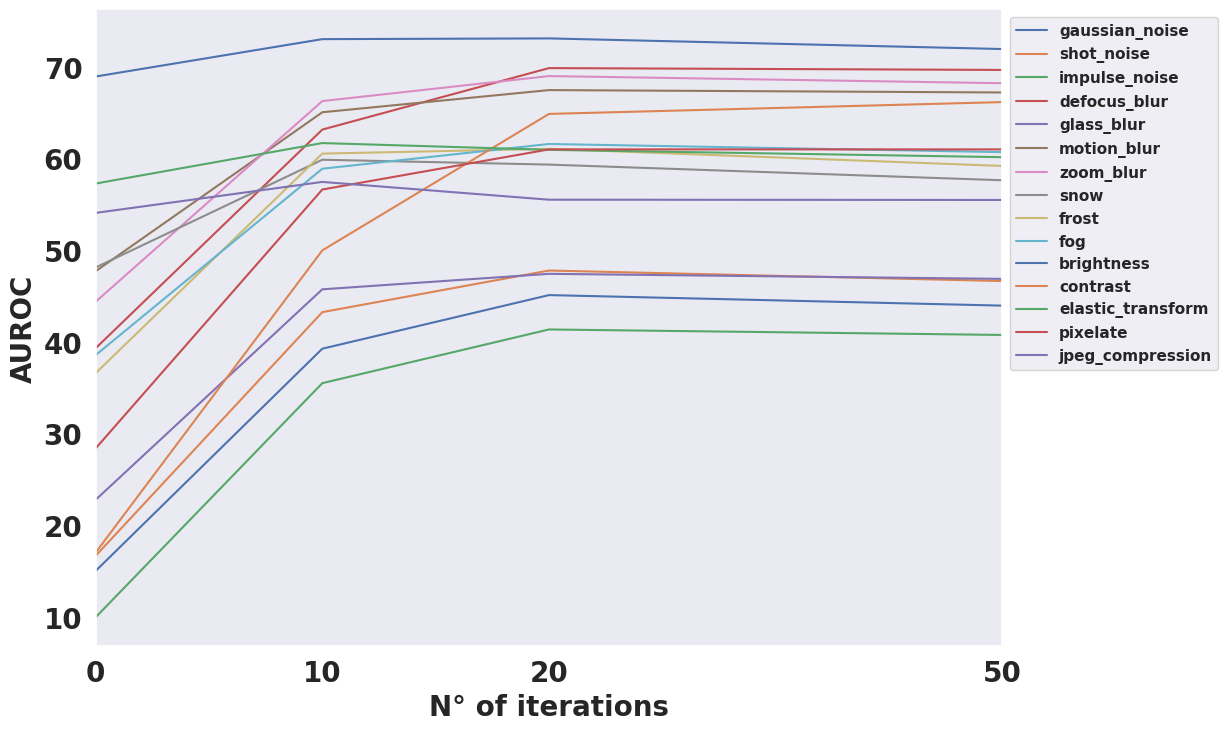}
    \caption{Performance (AUROC) reached by our method with different numbers of adaptation iterations on CIFAR-100C.}
    \label{fig:iterations_100}
\end{figure*}

Similarly to what was identified in previous studies~\cite{osowiechi2023tttflow, hakim2023clust3, osowiechi2024nc} and was confirmed for CIFAR-10C, also in the case of CIFAR-100C the best results are obtained after 20 adaptation iterations, while for some perturbation the same results can be obtained also with less interaction, after 20 the results tend to remain invariant for all the different perturbations. Figure~\ref{fig:iterations_100} shows for all the corruption of CIFAR-100C the results obtained at different iterations.

%\begin{comment}

\subsection{TinyImagenet-C}

Table~\ref{tab:performance-comparison-tiny} reports the results obtained on TinyImagenet-C, a dataset of 10.000 images with the same 15 corruptions described for CIFAR10-C and CIFAR100-C, but with 200 classes. \RecTTT{} outperforms all the other methods also on this dataset, with a $2.46\%$ improvement compared to \NCTTT{}, the second-best-performing model.

\begin{table*}[h]
\centering
\setlength{\tabcolsep}{4.5pt}
%\resizebox{\textwidth}{!}{%
\begin{tabular}{l|ccccccccc}
\toprule
\bf Corruption Type & \bf ResNet50 & \bf PTBN & \bf TENT & \bf TIPI & \bf \bf ClusT3 & \bf \NCTTT & \bf \RecTTT{} \\
\midrule
Gaussian Noise & 13.20 & 30.46 & 31.03 & 32.22 & 32.65 & 31.92 & \textbf{34.87} \\
Shot Noise & 16.28 & 32.26 & 33.07 & 34.27 & 34.72 &  34.47 &  \textbf{36.60}\\
Impulse Noise & 7.49 & 20.80 & 21.87 & 23.04 & 22.78 &  22.78 & \textbf{26.09} \\
Defocus Blur & 16.71 & 33.09 & \textbf{34.20} & 31.98 & 29.08&  25.28 & 31.09 \\
Glass Blur & 7.42 & 15.97 & 16.88 & 17.60 & 16.26 &  15.67 & \textbf{19.59} \\
Motion Blur & 27.71 & 43.09 & 44.40 & 43.54 & 43.92&  43.39 & \textbf{45.55} \\
Zoom Blur & 20.98 & 39.76 & 40.89 & 40.01 & 41.17 &  40.46 & \textbf{42.53}\\
Snow & 31.00 & 36.94 & 37.39 & 38.18 & 42.97 & \textbf{43.46} & 40.33 \\
Frost & 36.28 & 39.29 & 40.21 & 41.43 & 45.32 & \textbf{45.51} & 44.59\\
Fog & 16.40 & 31.51 & 32.52 & 32.82 & \textbf{37.85} & 37.68 & 33.08 \\
Brightness & 36.48 & 44.70 & 45.09 & 46.39 & \textbf{51.19} &  50.62 & 48.53\\
Contrast & 2.59 & 12.22 & \textbf{12.91} & 10.71 & 2.27 &  2.27 & 8.32 \\
Elastic Transform & 28.93 & 39.42 & 39.83 & 40.68 & 41.60 &  41.47 & \textbf{44.91} \\
Pixelate & 37.00 & 47.78 & 48.50 & 48.95 & 37.00 & 39.31 & \textbf{52.96}\\
JPEG Compression & 47.04 & 47.78 & 40.88 & 50.21 & 50.57 &  50.91 & \textbf{53.32} \\
\midrule
\textBF{Average} & 23.03 & 34.47 & 35.15 & 35.47 & 35.32 & 35.03 & \textbf{37.49}\\
\bottomrule
\end{tabular}
%} % end of \resizebox
\caption{Performance comparison with state-of-the-art on TinyImageNet-C perturbations (\%).}
\label{tab:performance-comparison-tiny}
\end{table*}

\section{On the contrastive loss performances}
\label{sec:cont}
To show the impact of our contrastive approach we implemented a TTT method based on the \textit{SimSiam}~\cite{chen2021exploring} framework. This solution only compares the features at the bottleneck level and is based on a single encoder, followed by a projection head and a predictor. The model was trained with the Cross-Entropy loss and the \textit{SimSiam} loss as auxiliary task.
As reported in the paper presenting the \textit{SimSiam} technique~\cite{chen2021exploring}, the loss is computed as the negative cosine similarity between \emph{i}) the features of the projector ($f_{E}$) extracted by the original image and \emph{ii}) the features of the predictor ($f_{P}$) of the augmented version of the image with a stop gradient on the predictor features.
To have a fair comparison with \RecTTT{}, we also used horizontal flip as augmentation.
During the adaptation phase, we adopted the same auxiliary loss to adapt the encoder features for a total of 20 iterations.

Table~\ref{tab:simsiam} shows that the \textit{SimSiam} contrastive learning approach, although achieving some good adaptation performances, does not achieve the same results as \RecTTT{}.
A possible reason for this result is that \textit{SimSiam} cannot fully capture the domain shift, which is hidden in the whole representation and not only at the bottleneck level. This is the main difference with \RecTTT{} that instead compares features at different layers.

\begin{table}[t!]
\centering
\setlength{\tabcolsep}{5pt}
%\resizebox{\linewidth}{!}{%
\begin{tabular}{l|c|c|c|c}
\toprule  
         & Impulse Noise & Brightness & Pixelate & Average \\
\midrule
SimSiam & 56.40 & 82.92 & 68.69 & 69.77\\
\midrule
\RecTTT{} & 69.28 & 94.03 & 82.13 & \textBF{82.82}\\
\bottomrule
\end{tabular}
%}
\caption{\textbf{On the contrastive loss.} Qualitative results using SimSiam contrastive approach on CIFAR-10C (\%).}
\label{tab:simsiam}
\end{table}

%Bibliography
\bibliographystyle{unsrt}  
\bibliography{paper}  

\end{document}